\PassOptionsToPackage{hyphens}{url}
\documentclass[sigconf, screen]{acmart}
\setcopyright{none}
\renewcommand\footnotetextcopyrightpermission[1]{}

\usepackage{amsmath} %
\usepackage{algorithmic}
\usepackage{graphicx}
\usepackage{textcomp}
\usepackage{xcolor}
\usepackage{booktabs}
\usepackage{enumitem}\setlist{nosep,leftmargin=*}

\newif\ifdraft\draftfalse

\newcommand{\sysname}{DuoArchitect}  %
\newcommand{\basesys}{DuoDiscover}  %
\usepackage{xspace}
\newcommand{\runLM}{DuoArch-Micro (Warm)\xspace}
\newcommand{\runLS}{DuoArch-Micro\xspace}
\newcommand{\runBS}{DuoArch-Nano\xspace}

\newif\ifshowcomments\showcommentsfalse
\newcommand{\newcommenter}[3]{\expandafter\newcommand\csname #1\endcsname[1]{%
  \ifshowcomments{\color{#3}\textsf{\scriptsize /*~#2:~##1~*/}}\fi}}
\usepackage{pgfplots}\pgfplotsset{compat=1.16}\usepgfplotslibrary{groupplots}\usetikzlibrary{positioning,fit,backgrounds,arrows.meta}
\newcommand{\stepnum}[1]{\tikz[baseline=(sn.base)]\node[circle,fill=black,text=white,inner sep=0.6pt,minimum size=1.15em,font=\scriptsize\bfseries](sn){#1};}
\usepackage{multirow}
\usepackage{needspace}
\usepackage{balance}
\let\oldsection\section
\let\oldsubsection\subsection
\def\BibTeX{{\rm B\kern-.05em{\sc i\kern-.025em b}\kern-.08em
    T\kern-.1667em\lower.7ex\hbox{E}\kern-.125emX}}
\AddToHook{para/begin}{\looseness=-1}
\begin{document}

\title[Discovering Bandwidth-Efficient Prefetchers to Shrink the Cache Hierarchy]{Discovering Bandwidth-Efficient Prefetchers\texorpdfstring{\\}{} to Shrink the Cache Hierarchy} %
\author{Patrick Tser Jern Kon}
\authornote{Both authors contributed equally to this work.}
\email{konp@google.com}
\affiliation{\institution{Google}\country{USA}}
\author{Akanksha Jain}
\authornotemark[1]
\email{avjain@google.com}
\affiliation{\institution{Google}\country{USA}}
\author{Abraham Gonzalez}
\email{abegonzalez@google.com}
\affiliation{\institution{Google}\country{USA}}
\author{Rui Meng}
\email{rmeng@google.com}
\affiliation{\institution{Google}\country{USA}}
\author{Dilara Soylu}
\email{soylu@stanford.edu}
\affiliation{\institution{Stanford University}\country{USA}}
\author{Shubham Agarwal}
\email{shubham3@berkeley.edu}
\affiliation{\institution{UC Berkeley}\country{USA}}
\author{Yawen Wang}
\email{yawenw@google.com}
\affiliation{\institution{Google}\country{USA}}
\author{Iman Hosseini}
\email{imanhosseini@google.com}
\affiliation{\institution{Google}\country{UK}}
\author{Hanna Alam}
\email{hannaalam@google.com}
\affiliation{\institution{Google}\country{USA}}
\author{Jinsung Yoon}
\email{jinsungyoon@google.com}
\affiliation{\institution{Google}\country{USA}}
\author{Tomas Pfister}
\email{tpfister@google.com}
\affiliation{\institution{Google}\country{USA}}
\author{Nanyun Peng}
\email{violetpeng@google.com}
\affiliation{\institution{Google}\country{USA}}
\author{Chun-Liang Li}
\email{chunliang@google.com}
\affiliation{\institution{Google}\country{USA}}
\author{Parthasarathy Ranganathan}
\email{parthas@google.com}
\affiliation{\institution{Google}\country{USA}}
\renewcommand{\shortauthors}{Kon, Jain, et al.}

\begin{abstract}
Shared last-level caches (LLCs) increasingly dominate server processor die area, forcing architects to trade core counts for cache capacity. While intelligent data prefetching offers a pathway to reclaim this silicon, designing coordinated, multi-level prefetching policies for multi-core systems is severely bottlenecked by day-long, cycle-level simulations and complex workloads. This simulation wall fundamentally limits even emerging LLM-based microarchitecture discovery frameworks; for instance, the state-of-the-art ArchAgent system evolves one cache level at a time on single-core benchmarks, adds multi-core configurations only at the end, and trails the human champion under multi-core shared cache contention.

To overcome this bottleneck, we introduce DuoArchitect, an autonomous agentic framework designed for multi-fidelity, multi-core co-discovery of prefetchers. %
In DuoArchitect, agents jointly evolve evaluation proxies and coordinated prefetching policies across the L1D, L2, and LLC hierarchies, constructing simulation proxies 3--26$\times$ faster than cluster evaluation and hardware diagnostic oracles to guide the search.
To study how much cache silicon a smarter prefetcher can replace, we evaluate and evolve prefetchers under reduced cache budgets, and find that evolving under our tightest cache constraint (with both the L2 and LLC halved) produced our strongest prefetcher overall at every cache size we tested: DuoArch-Nano. On the Fourth Data Prefetching Championship (DPC4) benchmarks with full-size caches, DuoArch-Nano delivers a 9.2\% geomean IPC speedup (+10.5\% on four-core mixes) while reducing four-core DRAM traffic by 11.4\%, more than doubling the gains of the human champion BertiGO and the agentic ArchAgent v2. When the LLC is halved, DuoArch-Nano erases the baseline's 8.6\% overall slowdown and edges out the full-LLC baseline overall (+0.88\%), demonstrating that such agentic frameworks can successfully 
generate prefetchers that reclaim megabytes of LLC SRAM per core with zero overall IPC penalty.

\end{abstract}

\maketitle
\pagestyle{plain} %

\section{Introduction}
\label{sec:intro}

Over the past decade, last-level cache (LLC) capacity on server system-on-chips (SoCs) has grown rapidly to keep pace with demand, consuming an increasingly massive share of valuable compute silicon. For example, the shared 32\,MB L3 takes roughly 44\% of the eight-core AMD Zen~4 core complex and 37\% of the full chiplet~\cite{munger2023zen, burd2024zen}.  This heavy reliance on large physical caches has run headfirst into a critical scaling barrier: while logic density continues to improve with each new technology node, high-density SRAM scaling has largely stalled~\cite{yeap20195nm, wu20223nm}. Consequently, server architects face a painful trade-off between cores and cache. They must either limit core counts to sustain sufficient cache capacity per core, or pack in more processing cores at the cost of drastically slashing the cache available to each~\cite{burd2024zen, aws2026graviton}.

\begin{figure}[t]
\centering
\includegraphics[width=\columnwidth]{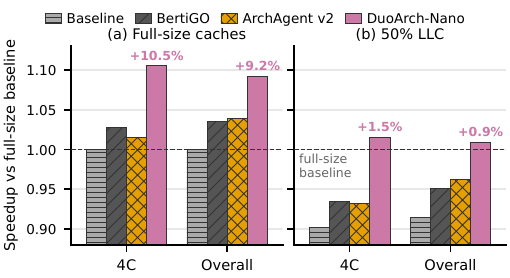}
\caption{Speedup over the full-size baseline, (a) with full-size caches and (b) with 50\% LLC.}
\vspace{-15pt}
\label{fig:intro}
\end{figure}

This silicon tax raises a fundamental architectural question: how much on-chip cache capacity can a smarter data prefetcher replace without losing performance? In principle, a prefetcher can hide off-chip memory latency and match the performance of a much larger LLC. In practice, building one for a modern cache hierarchy without increasing bandwidth requirements is exceedingly hard~\cite{srinath2007feedback,bera2021pythia}. %
A competitive prefetcher must coordinate across a multi-level cache hierarchy under strict metadata budgets~\cite{triage, ipcp}.
Furthermore, it must not degrade performance when LLC capacity and DRAM bandwidth are scarce~\cite{wu2011pacman, seshadri2015mitigating} and saturating off-chip memory buses~\cite{lee2008prefetch}. %

Autonomous LLM agents offer a promising way to navigate the vast design space of multi-level data prefetchers. Recently, ArchAgent v2~\cite{gonzalez2026archagentv2} showed that coding agents can evolve prefetchers that beat the championship-winning, hand-designed BertiGO~\cite{singh2026pushing} on single-core benchmarks.
However, agentic hardware design hits a brutal bottleneck: cycle-level simulation is orders of magnitude slower than software benchmarks, with a full four-core (4C) evaluation taking over a day per design. 
This simulation wall severely constrains the search space that prior agentic systems can explore.
ArchAgent v2 evolved only one cache level at a time on single-core and introduced multi-core configurations only in its final stage to save on simulation time.
This cascaded approach leaves its design about 1.3\% behind BertiGO under 4C shared-LLC contention while using 7\% more 4C DRAM traffic (14\% more than the baseline).

To scale agentic discovery beyond this simulation wall for multi-core systems, we introduce \textbf{\sysname{}}. Building on the multi-fidelity paradigm of \basesys{}~\cite{agarwal2026duodiscover}, \sysname{} dispatches fleets of sandboxed agents that co-evolve all three cache levels at once. Rather than waiting over a day for full simulation at every iteration, agents co-evolve prefetcher designs and fast local evaluators, called \textit{yardsticks}, inside their sandboxes, substantially accelerating iterations while enabling joint multi-core optimization across all three cache levels. Promising candidates are then evaluated on a disclosed search set across a distributed simulation cluster before final evaluation on the full official set, which adds held-out workloads.

We apply \sysname{} to the Fourth Data Prefetching Championship (DPC4) setup~\cite{dpc4}, achieving state-of-the-art results across all regimes (\S\ref{sec:b50results}) and outperforming existing works by large margins. Over a $\sim$13-day search, our flagship run (\textbf{\runBS{}}) screened an estimated $\sim$19,800 candidates on local yardsticks, evaluated 1,521 on the search set, and ran the official evaluation on only 22 new best designs. The resulting three-level prefetcher delivers a 9.2\% overall geomean IPC speedup with the full-size DPC4 caches (Fig.~\ref{fig:intro}a), with 9.6\% at full bandwidth, 7.5\% at limited bandwidth, and 10.5\% on 4C (Fig.~\ref{fig:settings_all}). This more than doubles the gains of both BertiGO (3.5\%) and ArchAgent~v2 (3.9\%). More importantly, these performance gains do not come at the cost of extra memory traffic. In fact, \runBS{} reduces 4C DRAM traffic by 11.4\% compared to the baseline.

\noindent\textbf{Key Findings.}
Mirroring the reduced per-core LLC of cloud-dense server SoCs~\cite{burd2024zen, aws2026graviton},
we also explore whether fast multi-fidelity search can trade LLC silicon for prefetcher intelligence by evaluating and evolving prefetchers under reduced cache budgets (\S\ref{sec:b50results}, \S\ref{sec:scaling}), uncovering two central findings:
\begin{itemize}
\item \textbf{Prefetcher intelligence can replace half the LLC.} Cutting the LLC by 50\% (from 3\,MB to 1.5\,MB per core) slows the baseline down by 8.6\%. \runBS{} erases this entire penalty and edges out the full-LLC baseline overall (+0.88\%; Fig.~\ref{fig:intro}b), reclaiming 1.5\,MB of SRAM per core with no overall IPC loss. The key reason behind \runBS{}'s competitiveness even with half the LLC size is its high accuracy and bandwidth efficiency. In particular, \runBS{} uses 5--8\% less DRAM traffic than the full-LLC baseline on single-core workloads and matches its four-core DRAM traffic (within 0.4\%).
\item \textbf{Cache scarcity in evolution breeds stronger prefetchers.} \runBS{} was evolved under our tightest cache constraint, yet it beats prefetchers evolved on larger caches overall on every machine size, including with the full-size 12\,MB LLC (+9.2\% vs.\ 7.1--8.0\%). This suggests that generous LLCs mask prefetch pollution, whereas severe cache scarcity acts as a regularizer that forces strict efficiency and fairness controls.
\end{itemize}

\noindent\textbf{Microarchitectural Insights.}
Inspecting \runBS{} reveals that performance gains come from a combination of diverse prefetch engines, along with clever coordination and cache management. We identify three insights behind the performance gains (\S\ref{sec:both50design}, \S\ref{sec:ablations}):

\begin{itemize}
\item \textbf{Diverse Prefetch Engines.} Across the three cache levels, \runBS{} employs a diversity of prefetch engines spanning strided, spatial, and temporal prefetchers. The prefetchers are modulated based on congestion feedback, and cover both global and local patterns.
\item \textbf{Communication across levels.} Rather than acting in isolation, all three levels of prefetchers coordinate. 
\item \textbf{Cache management.} The design of \runBS{} recognizes that prefetchers cannot be effective without co-designed cache management. Since the DPC setup doesn't allow modification of replacement state directly, prefetch engines indirectly modulate replacement priority for slow cores or critical lines (such as PTE entries) by issuing prefetch requests to cache-resident lines.
\end{itemize}

\noindent\textbf{Contributions.}
This paper makes the following contributions: 

\begin{itemize}
\item \textbf{Scaling multi-fidelity agentic discovery to cycle-level simulation.} We present \sysname{}, a system using agent-built fast ranking proxies and diagnostic hardware oracles to co-evolve three-level prefetchers (\S\ref{sec:method}, \S\ref{sec:b50found}).
\item \textbf{A new state-of-the-art three-level prefetcher.} \runBS{} achieves a 9.2\% overall IPC speedup on DPC4 (+10.5\% on 4C) while cutting 4C DRAM traffic by 11.4\%, more than doubling prior human and agentic speedups (\S\ref{sec:b50results}, \S\ref{sec:both50design}).
\item \textbf{Trading LLC capacity for prefetcher intelligence.} We show that an evolved prefetcher compensates for halving the LLC overall (+0.88\% vs.\ the full-LLC baseline), reclaiming 1.5\,MB of LLC SRAM per core with no overall IPC loss (\S\ref{sec:b50results}).
\item \textbf{Demonstrating that cache scarcity breeds stronger prefetchers.} Across evolutionary runs on reduced-cache processors, we find that the tightest cache constraint produced retention and shared-LLC QoS mechanisms that generalize best overall across the cache sizes we tested (\S\ref{sec:scaling}).
\end{itemize}

\section{\sysname{}}
\label{sec:method}

\begin{figure}[t]
\centering
\includegraphics[width=\columnwidth]{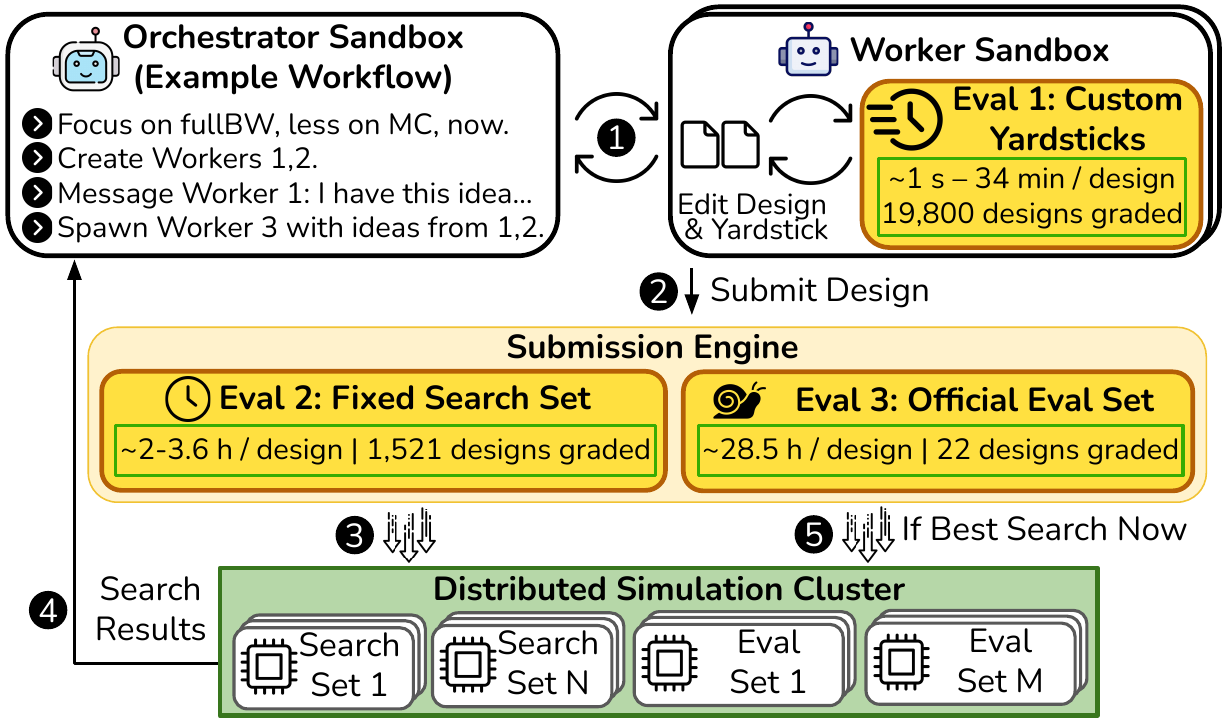}
\caption{Overview of \sysname{} and its 3 evaluation tiers.}
\label{fig:overview}
\end{figure}

\sysname{} builds on \basesys{}~\cite{agarwal2026duodiscover}, a multi-fidelity framework for autonomous agentic discovery, and adapts it for hardware simulation driven feedback (Fig.~\ref{fig:overview}). Because a full DPC4 (details in \S\ref{sec:methodology}) evaluation takes over a day, one to two orders of magnitude slower than the software benchmarks targeted by \basesys{}, \sysname{} separates evaluation into a three-tier funnel.
\stepnum{1} Within each fleet, an \textbf{Orchestrator} steers exploration by deciding on the objectives to pursue and acting on them by spawning, communicating with, and merging prefetcher designs across parallel \textbf{Workers}. Inside its sandbox, each worker iteratively edits the three-level prefetcher design, and fast evaluators (which we call \textbf{Yardsticks}, $\sim$1\,s--34\,min per design) to filter out weak ideas before \stepnum{2} submitting a promising candidate to the \textbf{Submission Engine}. The engine \stepnum{3} schedules evaluation on the \textbf{Fixed Search Set} (131 disclosed single-core traces and 14--15 4C mixes at 5M/20M from DPC4 at $\sim$2--3.6\,h per design) across the \textbf{Distributed Simulation Cluster} and \stepnum{4} returns the verified search results to the orchestrator. Finally, whenever search produces a new best design, it is \stepnum{5} evaluated asynchronously offline on the \textbf{Official Evaluation Set}: all 610 single-core traces at 50M/200M and 483 4C mixes at 20M/50M per core from DPC4 (25--28.5\,h per design), comprising the 131 traces and 14--15 mixes from the search set plus 479 traces and 468--469 mixes held out to measure unseen generalization. Below, we detail how we build on \basesys{} across this pipeline.

\noindent\textbf{Orchestrator--worker loop.} Each worker sandbox is initialized with the three-level C++ prefetcher seed within strict storage budgets, read-only access to the 131 disclosed traces, and a parallelized custom yardstick template that workers freely customize to screen candidate edits locally before submitting (\S\ref{sec:yardsticks}). Each sandbox also holds a writable private copy of the ChampSim simulator source, allowing workers to instrument the simulator itself into custom diagnostic oracles (\S\ref{sec:b50found}).
Adapting base \basesys{} required $\sim$1.8k lines of framework changes across four mechanisms: (1) a read-only Docker tool sandbox for the orchestrator that restricts its view to the run directory so it cannot inspect offline official scores or copy ungraded code between workers; (2) a multi-objective ledger that stores fullBW, limitBW, and 4C speedups separately, allowing the orchestrator to dynamically adjust objective weights 
and re-rank all past designs without re-simulating; (3) a sub-percent deterministic noise gate replacing base \basesys{}'s 1\% statistical noise floor so deterministic 0.1--3\% IPC gains are preserved rather than discarded as noise; and (4) turn and budget guardrails. The guardrails cap worker turns at 2 hours and give workers that merge other workers' designs first claim on free worker slots and the grade queue. They also block new turns or design merges when less time remains in a step than a grade requires. Across 24-hour steps, the framework auto-generates step briefings strictly from disclosed search-set results, with an automated check stripping any official scores from prompts and notes.

\noindent\textbf{Submission Engine.} 
Candidate design submissions by workers \stepnum{2} are handed to a Submission Engine.
To prevent wasted or corrupted multi-hour cluster runs, the engine freezes a snapshot on arrival (so subsequent worker edits cannot alter a queued submission), limits each worker to one evaluation in flight, and immediately returns stored scores if an identical design and machine configuration was already evaluated. For each new submission, the engine compiles and checks the snapshot on a clean copy of the DPC4 simulator inside a network-less container through four legality gates: (1) a fail-closed source check rejecting access to private cache state (the cache's tag and MSHR arrays) or external includes, (2) a fail-closed build check rejecting forbidden file, environment, process, or assembly calls, (3) a fail-closed storage probe enforcing the 32\,KB L1D, 128\,KB L2, and 256\,KB LLC metadata budgets, and (4) an exact-match plagiarism check against the 8 published DPC4 submissions. Passing builds \stepnum{3} fan out across the simulation cluster as $\sim$277 tasks covering all 131 disclosed traces (at fullBW and limitBW) and 14--15 disclosed 4C mixes.
To make search-set evaluation an order of magnitude faster than a 50M/200M official evaluation without distorting candidate rankings, the search set shortens the simulation window by $10\times$ to 5M/20M instructions (15M/60M on Google traces, where short warmups mislead) rather than simulating fewer traces. In our calibration on the 8 published DPC4 submissions, shortening the window to 15M/60M preserved full-length rankings (Kendall's rank correlation $\tau = 0.857$~\cite{kendall1938new}, where 1 indicates identical order and 0 indicates no correlation) and 5M/20M matched the 15M/60M ordering ($\tau = 1.00$), whereas running a smaller subset of traces flipped the winner ($\tau = 0.50$). Every baseline IPC is frozen per cache configuration and instruction window (5M/20M for search; 50M/200M single-core and 20M/50M per core on 4C for official evaluation) so speedups always compare matching cache sizes and windows, while cluster array tasks enforce per-site memory reservations with automatic retry for out-of-memory kills and stragglers. If any trace or mix crashes, the entire grade fails; otherwise, the engine \stepnum{4} returns the per-part speedups to the ledger (while also re-running the worker's yardstick on the frozen snapshot to log its agreement $\tau$).

\section{\runBS{} Design}
\label{sec:both50}
\label{sec:designs}
\label{sec:both50design}

\begin{figure}[t]
\centering
\includegraphics{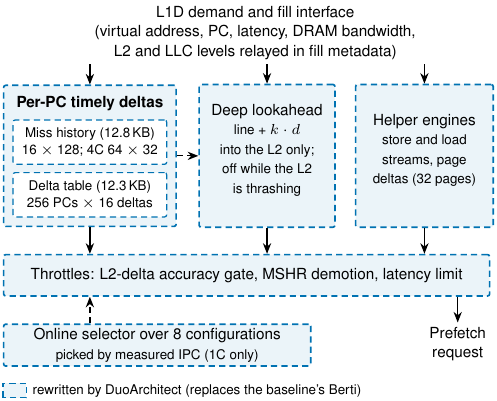}
\caption{L1D of \runBS{}: stages deep lookahead into L2, with stream and page-delta helpers and an online configuration selector.}
\label{fig:b50_l1d}
\end{figure}

\begin{figure}[t]
\centering
\includegraphics{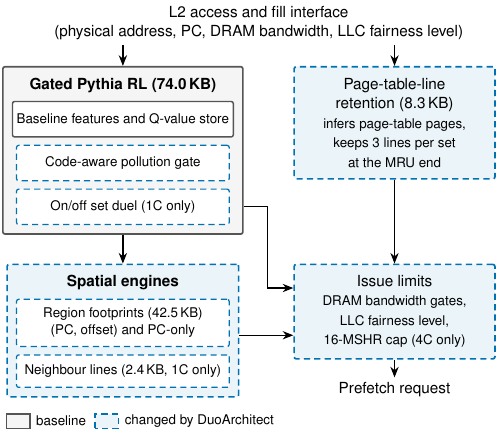}
\caption{L2 of \runBS{}: Pythia behind a code-aware gate, with region-footprint and neighbour engines and a page-table-line retention engine.}
\label{fig:b50_l2}
\end{figure}

\begin{figure}[t]
\centering
\includegraphics[width=\columnwidth]{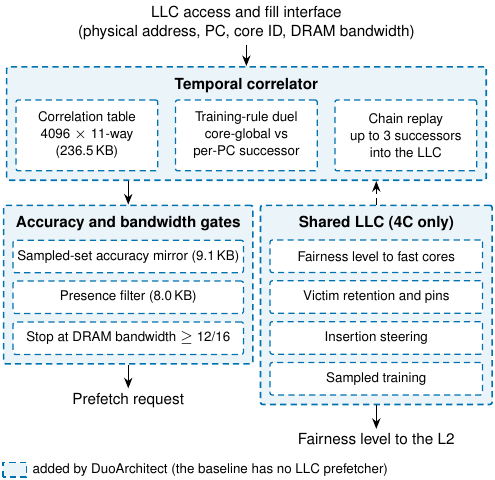}
\caption{LLC of \runBS{}: a temporal correlator with accuracy and bandwidth gates and four-core sharing controls.}
\label{fig:b50_llc}
\end{figure}

Figs.~\ref{fig:b50_l1d}--\ref{fig:b50_llc} illustrate the three-level architecture of \runBS{} (utilizing 99.2\%, 99.4\%, and 99.8\% of the L1D, L2, and LLC DPC4 storage budgets), showing how each cache level pairs stronger pattern coverage with explicit cache pollution, and bandwidth controls.

\noindent\textbf{L1D (Per-PC Timely Deltas and L2 Look-Ahead Staging).} While the baseline's Berti prefetcher (\S\ref{sec:methodology}) is keyed by page, a single page often mixes unrelated load instructions with different strides. \runBS{} therefore re-keys Berti by load PC (Fig.~\ref{fig:b50_l1d}), tracking per-instruction miss timestamps to learn timely address deltas and using each delta's historical coverage to decide whether prefetches fill directly into the L1D or only into the L2. The highest-confidence delta stages deeper look-ahead steps into the L2, so subsequent L1D prefetches hit in the L2. This look-ahead automatically pauses whenever the L2 reports heavy prefetch pollution or the LLC relays a 4C fairness backoff signal. 

To capture access patterns that a single-PC delta table misses, two helper engines run alongside Berti: (1) a stream engine that predicts store misses from L1D fills that were not triggered by any prior load or prefetch, (2) a page-delta engine that tracks spatial strides across loops with multiple load PCs. The stream engine pauses on low accuracy or high DRAM bandwidth use; the page-delta engine pauses on high DRAM bandwidth use.

All L1D engines operate on virtual addresses, and do not stop at a page boundary, so deltas and deep look-ahead can cross into the next page. Under congestion, L1D prefetches demote to L2 fills under high MSHR occupancy ($\ge$10 of 16 MSHRs busy) and throttle to one issue per trigger when average demand-miss latency exceeds 350 cycles. Prefetch degree and lookahead are dynamically determined by an online epoch arbiter that periodically tests alternative degree and look-ahead depths, switching only when a configuration improves measured IPC.

\noindent\textbf{L2 (Pollution-Gated Pythia, Spatial Footprint Engine, and PTE Retention).} At the halved 1\,MB L2 (Fig.~\ref{fig:b50_l2}), Pythia's aggressive RL-driven requests frequently evict useful instructions and data or saturate limited DRAM bandwidth. \runBS{} places Pythia behind a 36.5\,KB shadow-tag monitor (2,048 sets $\times$ 8 ways of 10-bit partial tags within the L2 prefetcher budget) that tracks L2 data evictions and detects when Pythia's fills displace instruction lines (inferred when a fill evicts an address absent from the data-only shadow) or re-demanded data. \runBS{} throttles Pythia to a single prefetch per access under moderate pollution ($>$50\% harmful-to-useful ratio) and to a 1-in-8 rate under severe pollution ($>$90\%, retaining enough prefetches to detect when Pythia becomes useful again, complemented on 1C by a set-dueling on/off switch and stricter accuracy gates under high bus utilization). 

To cover spatial access patterns that Pythia misses or throttles back on, two spatial predictors are added: (1) a region-footprint engine~\cite{sms} that records and replays bitvectors of accessed lines within 2\,KB regions, and (2) a neighbour engine that learns per load PC which nearby line offsets ($\pm$1 to $\pm$4 lines) miss soon after on at least half of its misses and prefetches them.

Because halving both caches causes leaf page-table (PTE) lines to be evicted between address translations, the L2 introduces an 8.3\,KB \textbf{page-table-line retention engine}, which identifies PTE lines from recurring page-walk fills (lines filled into the L2 that no prior load or prefetch requested and no data access ever touches) and marks up to three recurring PTE lines per L2 set in the shadow tags. On every set access, the engine issues a prefetch to each marked line (generating an L2 hit that refreshes it to the MRU position without memory traffic). Across all L2 engines, issue rates are governed by real-time DRAM bandwidth thresholds, the LLC's 4C fairness signal, and an MSHR occupancy cap on four cores.

\noindent\textbf{LLC (Temporal Correlator and Four-Core QoS Coordinator).} Because the baseline leaves the 256\,KB LLC prefetcher slot empty, irregular pointer-chasing misses that escape the L1D and L2 go unprefetched. \runBS{} fills this slot with an on-chip temporal (Markov) address-correlation table that links each LLC miss to its successor and replays chains of up to three successors upon a repeat miss (Fig.~\ref{fig:b50_llc})~\cite{triage}.  The engine uses set-dueling to choose between per-core and per-PC successor training. Prefetch issue is gated on DRAM bus utilization and a sampled accuracy mirror, which tracks whether prefetched lines are used on 1 in 16 sampled LLC sets. 

Under four-core execution (4C), where all four cores contend for the halved 6\,MB LLC, the LLC prefetcher also serves as a central QoS coordinator to protect the slowest core in a mix from co-runner thrashing in four ways. First, whenever its 64K-cycle check detects a memory-bound, irregular victim core ($<$0.5 IPC, $\ge$5 L2 demand MPKI reaching the LLC, and $<$2 L2 prefetches per demand miss), it relays a backoff signal through the L2 to silence fast aggressor cores' L1D prefetchers. Second, it repurposes two of the correlation table's 11 ways on 4C to \textbf{pin one LLC-hit line per set for the slowest victim core ($<$0.5 IPC)}: whenever a co-runner fills into that set, the prefetcher issues a prefetch to the stored victim address, generating an LLC hit that refreshes its replacement age. Third, for streaming co-runners ($<$15\% LLC hit rate), it steers DRRIP's set-dueling selector~\cite{jaleel2010rrip} toward distant (evict-soon) insertion by answering their fills into distant-insertion leader sets with a prefetch into near-insertion leader sets. Finally, it restricts cores whose miss sequences rarely repeat ($<$1/16 of writes matching an existing entry or $<$31\% accuracy) to 1-in-8 writes into the shared correlation table so they cannot overwrite the victim core's learned chains.
\section{Methodology}
\label{sec:methodology}
\label{sec:background}

\begin{table}[t]
\centering\footnotesize
\caption{DPC4 processor hierarchy and evaluation regimes.}
\label{tab:dpc4}
\setlength{\tabcolsep}{4pt}
\begin{tabular}{@{}lp{0.72\columnwidth}@{}}
\toprule
Component & Configuration \\
\midrule
L1D cache & 48\,KB/core, 12-way, 5 cycles, 16 MSHRs; Berti baseline, 32\,KB PF budget \\
L2 cache & 2\,MB/core, 16-way, 10 cycles, 32 MSHRs; Pythia baseline, 128\,KB PF budget \\
Shared LLC & 3\,MB/core, 12-way, 35 cycles, 64 MSHRs/core, DRRIP; no baseline prefetcher, 256\,KB PF budget \\
DRAM & 1 channel, 4,800\,MT/s (fullBW, 4C) or 800\,MT/s (limitBW)\\
1C regimes & 610 traces (50M/200M) at 4,800\,MT/s (fullBW) and 800\,MT/s (limitBW) \\
4C regime & 483 4-trace mixes (20M/50M per core) at 4,800\,MT/s, harmonic-mean IPC \\
\bottomrule
\end{tabular}
\end{table}

\noindent\textbf{The DPC4 benchmark.} DPC4 evaluates three-level data prefetchers in ChampSim~\cite{gober2022championship} across two single-core bandwidth regimes (\textbf{fullBW} and \textbf{limitBW}) and a four-core shared-LLC regime (\textbf{4C}), summarized in Table~\ref{tab:dpc4} (with 4C evaluated across 483 of the 484 mixes, excluding one Graph mix where the baseline deadlocks). The benchmark spans four workload suites: SPEC CPU 2017, Google server, AI/ML, and Graph. To prevent overfitting during design, 131 single-core traces and 14--15 4C mixes are disclosed for search, while the remaining 479 traces and 468--469 mixes are held out strictly for final evaluation.

\noindent\textbf{Baseline and prior prefetchers.} The DPC4 \textbf{baseline} pairs Berti at the L1D~\cite{navarro2022berti} with Pythia at the L2~\cite{bera2021pythia} and leaves the 256\,KB LLC slot empty. Berti timestamps L1D misses and fills to measure fetch latency and learn high-coverage timely address deltas, selecting L1D or L2 fill placement from coverage and MSHR occupancy (though ChampSim's DPC4 baseline indexes Berti by page rather than load PC, mixing unrelated instructions). Pythia formulates L2 prefetching as online tabular reinforcement learning (SARSA), mapping context features (PC+delta and recent delta sequences) to intra-page offset actions and penalizing inaccurate prefetches more heavily when memory bandwidth is busy; however, without a confidence gate, it still issues low-accuracy prefetches on irregular workloads (45--55\% accuracy on DPC4, Table~\ref{tab:mpki}). 

Apart from baseline, we compare against two strong prior three-level designs:
(1) \textbf{BertiGO}~\cite{singh2026pushing}, the DPC4 human winner, which pairs a per-PC timely-delta L1D prefetcher with a global stride tracker and spatial/delta tables at the L2; and (2) \textbf{ArchAgent~v2}~\cite{gonzalez2026archagentv2}, an evolutionary LLM agent system that evolved L1D (augmenting Berti with global stride and delta tables and an RL arbitrator), L2 (5-feature sum-pooled Pythia), and LLC (step-ratio temporal/stride prefetcher) sequentially in isolation. 

\noindent\textbf{Cache configurations and metrics.} We evaluate the DPC4 configuration with large caches (Table~\ref{tab:dpc4}; 2\,MB L2 and 3\,MB LLC per core) alongside two cache-constrained configurations. Both keep the 48\,KB L1D and all prefetcher storage budgets fixed: (1) 50\% LLC (2\,MB L2 and 1.5\,MB LLC per core, or 6\,MB shared across four cores, cutting 1.5\,MB of SRAM per core), and (2) 50\% L2 + 50\% LLC (1\,MB L2 and 1.5\,MB LLC per core, cutting 2.5\,MB of SRAM per core). Our flagship prefetcher, \runBS{}, is evolved from the unmodified baseline on the most constrained machine (with both the L2 and LLC halved); \S\ref{sec:both50design} presents its three-level architecture and \S\ref{sec:results} its performance, while \S\ref{sec:scaling} ablates why SRAM scarcity breeds higher performance by comparing against runs evolved at 50\% LLC.

We report official gains on two complementary scales. The first, \textit{vs.\ full-size baseline}, normalizes speedups against the baseline with full-size caches (so $\ge 0\%$ means a reduced-cache configuration matches or beats the full-cache baseline). The second, \textit{vs.\ same-size baseline}, normalizes speedups against the baseline on that same reduced-cache machine, measuring how much prefetching improves a given cache configuration.

\noindent\textbf{Experimental setup.} Across our evolutionary runs (\S\ref{sec:evaluation}), each run evolves prefetchers for $\sim$13 days in 24-hour steps using up to 8 parallel fleets of 4 workers on 44 CPU cores, with Claude Opus~5.5 for both the  workers and the orchestrator. Even with custom yardstick screening, each run grades 800--1,521 candidates on the search set (consuming 50,000--133,000 search core-hours per run) and each \stepnum{5} offline official evaluation 
costs $\sim$3,000 additional core-hours, totaling over 300,000 core-hours across our runs, making each evolutionary run a substantial cluster investment.
\section{Evaluation}
\label{sec:evaluation}
\label{sec:results}

Because L2 and LLC SRAM occupy a large fraction of processor die area, architects face a fundamental question: how much on-chip cache capacity can a smarter prefetcher replace without losing performance? Even more intriguingly, does evolving a prefetcher under severe cache scarcity force the discovery of stronger general-purpose mechanisms that excel even with larger cache sizes? 

\oldsubsection{Results Across Machine Sizes}
\label{sec:b50results}
\begin{figure*}[t]
\centering
\includegraphics[width=0.95\textwidth]{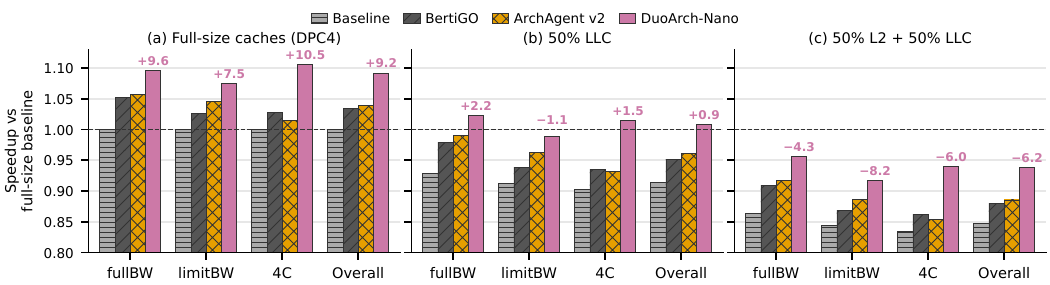}
\caption{Official speedup over the baseline with full-size caches for every setting on each machine.}
\label{fig:settings_all}
\end{figure*}

Fig.~\ref{fig:settings_all} evaluates the design of \runBS{} across all three processor configurations alongside the fixed baseline prefetcher, the DPC4 human champion (BertiGO), and the prior LLM-agent state of the art (ArchAgent~v2). Walking from full cache capacity down to the most constrained machine reveals three headline results:
\begin{enumerate}
\item \textbf{State-of-the-art performance with full-size caches:} Even though \runBS{} was evolved entirely on the smallest cache, running with the full-size DPC4 caches achieves a 9.21\% overall speedup over baseline (+9.58\% fullBW, +7.54\% limitBW, and +10.52\% 4C), more than doubling the gains of BertiGO (3.50\%) and ArchAgent~v2 (3.90\%; 3.8\% as reported in~\cite{gonzalez2026archagentv2}).
\item \textbf{Beating the full-LLC baseline with half the LLC:} Cutting 1.5\,MB of LLC per core slows the baseline processor down by $-$8.56\% ($-$7.07\% on fullBW, $-$8.80\% on limitBW, and $-$9.80\% on 4C). Fixed BertiGO and ArchAgent~v2 achieve slightly larger relative speedups over the same-size baseline than at full size (rising from 3.50\% to 4.11\% and from 3.90\% to 5.32\%), confirming that prefetching matters more when caches get smaller. However, their full-size tuning cannot overcome the missing LLC capacity under limited bandwidth and 4C contention, leaving them at $-$4.91\% and $-$3.82\% below the baseline with the full-size caches. Yet \runBS{} overcomes the missing LLC capacity and scores +0.88\% above the baseline with the full-size caches (+10.16\% over the same-size baseline), demonstrating that an evolved prefetcher can replace half the LLC without sacrificing overall IPC.
\item \textbf{Recovering most of the loss with caches halved:} When both the L2 and LLC are halved, the baseline processor drops by $-$15.28\% ($-$13.58\% / $-$15.66\% / $-$16.56\%), and fixed BertiGO and ArchAgent~v2 plunge to $-$12.06\% and $-$11.46\% vs.\ the full-size baseline. While halving both caches is too large a hardware penalty to erase completely, \runBS{} recovers 59\% of that official two-cache loss (climbing to $-$6.20\% vs.\ the full-size baseline) and achieves +10.52\% over the same-size baseline, outperforming BertiGO (+3.95\%) and ArchAgent~v2 (+4.66\%) on that machine by 5.9--6.6 percentage points. That a single prefetcher evolved under the tightest cache constraint dominates across all machines suggests that cache scarcity during evolution breeds stronger, more generalizable prefetchers (\S\ref{sec:scaling}).
\end{enumerate}
Across each evaluation regime, \runBS{} outperforms all prior designs on fullBW, limitBW, and 4C at every machine size.

\begin{table}[t]
\centering\footnotesize
\setlength{\tabcolsep}{3pt}
\caption{Cache counters on the full-size machine.}
\label{tab:mpki}
\begin{tabular}{@{}lrrrrrrr@{}}
\toprule
 & L1D & L2 & LLC & L1D & L2 & L2 PF & DRAM \\
Design & MPKI & MPKI & MPKI & Acc. \% & Acc. \% & PKI & PKI \\
\midrule
\multicolumn{8}{@{}l}{\emph{1C FullBW}} \\
Baseline & 22.33 & 10.09 & 6.53 & 70.7 & 47.2 & 49.0 & 21.0 \\
ArchAgent v2 & \textbf{18.36} & \textbf{8.14} & 5.56 & 81.1 & 67.3 & \textbf{35.4} & 18.5 \\
\runBS{} & 22.15 & 8.35 & \textbf{5.15} & \textbf{83.3} & \textbf{69.2} & 53.7 & \textbf{18.0} \\
\midrule
\multicolumn{8}{@{}l}{\emph{1C LimitBW}} \\
Baseline & 23.26 & 10.20 & 6.07 & 70.9 & 45.1 & 53.9 & 20.9 \\
ArchAgent v2 & \textbf{20.00} & \textbf{8.39} & 5.60 & 83.4 & 51.1 & \textbf{30.7} & 18.1 \\
\runBS{} & 23.79 & 10.26 & \textbf{5.52} & \textbf{83.6} & \textbf{76.1} & 50.0 & \textbf{17.4} \\
\midrule
\multicolumn{8}{@{}l}{\emph{4C}} \\
Baseline & 76.02 & 21.21 & 12.25 & 84.4 & 54.9 & 185.9 & 52.1 \\
ArchAgent v2 & \textbf{62.53} & 17.06 & 11.61 & 90.7 & \textbf{70.1} & 179.5 & 59.3 \\
\runBS{} & 71.53 & \textbf{16.24} & \textbf{8.50} & \textbf{91.3} & 66.8 & \textbf{110.1} & \textbf{46.1} \\
\bottomrule
\multicolumn{8}{@{}l}{\scriptsize MPKI counts demand misses. PF: L2 prefetches issued. DRAM: DRAM requests.}\\
\multicolumn{8}{@{}l}{\scriptsize PKI: per kilo-instruction. Means per trace (610) or 4C mix-core (483 mixes).}
\end{tabular}
\end{table}

\begin{figure}[t]
\centering
\includegraphics[width=\columnwidth]{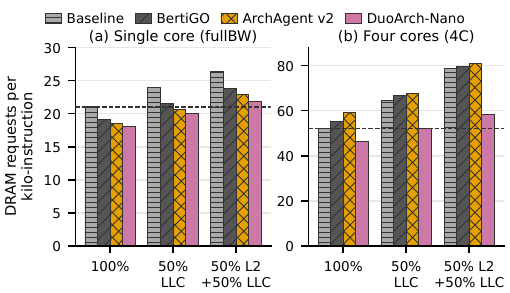}
\caption{DRAM requests per kilo-instruction per machine. The dashed line marks the baseline with full-size caches.}
\label{fig:dram_pki}
\end{figure}

\noindent\textbf{Where the speedup comes from.}
\label{sec:b50why}
Table~\ref{tab:mpki} compares demand misses, prefetch accuracy, L2 prefetch volume, and DRAM traffic of the baseline, ArchAgent~v2, and \runBS{} on the full-size machine. Across all three regimes, \runBS{} combines the lowest LLC demand misses with the highest L1D prefetch accuracy (83.3--91.3\%), high L2 accuracy (66.8--76.1\%), and the lowest DRAM traffic (17.4--46.1\,PKI). On four cores, whereas prior prefetchers issue many speculative L2 requests that contend for the shared LLC and memory bus (185.9 and 179.5\,PKI), \runBS{} stages 33.7\,PKI of high-accuracy L1D look-ahead into the L2 and cuts 4C L2 prefetches by 41\% (to 110.1\,PKI) while still reducing 4C LLC demand misses by 30.6\% (12.25~$\rightarrow$ 8.50\,MPKI). It achieves this selectivity by gating low-confidence speculation across all three levels and, when irregular accesses cannot be predicted accurately, retaining resident lines and steering streaming fills toward early eviction (\S\ref{sec:both50design}).

The four-core memory subsystem is already queue-bound at full cache size ($\sim$1,800-cycle average LLC miss latency vs.\ $\sim$200 cycles on a single core), yet every other evaluated prefetcher increases 4C DRAM traffic above the baseline (52.1\,PKI; Fig.~\ref{fig:dram_pki}), reaching 53.7\,PKI (+3\%) for \runLM{}, 55.4\,PKI (+6\%) for BertiGO and \runLS{}, and 59.3\,PKI (+14\%) for ArchAgent~v2. By contrast, \runBS{} is the only design that reduces 4C off-chip traffic below the full-size baseline, cutting DRAM requests by 11.4\% to 46.1\,PKI. Eliminating wasted off-chip traffic enables \runBS{} to replace physical cache capacity. Halving the shared LLC pushes baseline 4C DRAM traffic up by 24\% to 64.7\,PKI, and BertiGO and ArchAgent~v2 higher still (66.6 and 67.6\,PKI). \runBS{} instead holds 50\%-LLC 4C DRAM traffic to 52.3\,PKI, 19\% below the same-size baseline and within 0.4\% of the full-size baseline's 52.1\,PKI; with both caches halved, it stays 26\% below the same-size baseline (58.1\,PKI). This relieves memory-bus queueing, so \runBS{} beats the full-size baseline on 4C (+1.54\%) with half the LLC.

A production design could also reuse the saved cache silicon for additional cores or accelerators. For instance, based on the AMD Zen~4 core-complex area breakdown (where the 32\,MB L3 occupies $\sim$44\% of the eight-core complex~\cite{munger2023zen, burd2024zen}), halving the per-core LLC shrinks each core+cache tile by $\sim$16\%, freeing enough silicon area for roughly 20\% more cores.

\begin{figure*}[t]
\centering
\includegraphics[width=\textwidth]{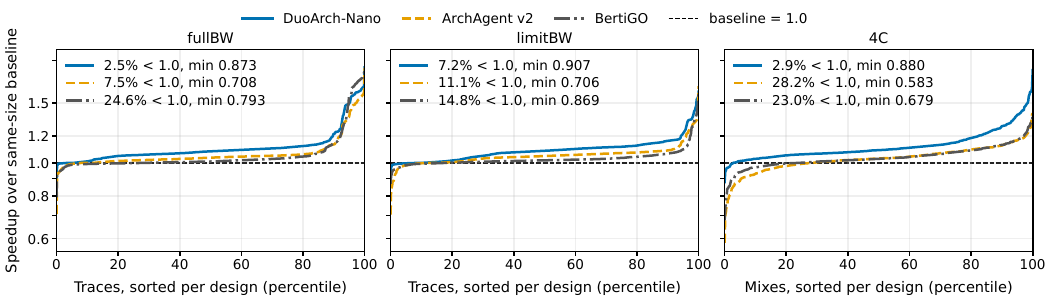}
\caption{Speedup over the same-size baseline on the 50\% L2 + 50\% LLC machine for every trace at fullBW/limitBW and every 4C mix, sorted per design.}
\label{fig:b50_scurve}
\end{figure*}

\noindent\textbf{Per-trace robustness and held-out generalization.} Does evolving under severe cache scarcity cause the prefetcher to overfit the  search traces or hurt outlier workloads when cache space is tight? 

The answer is no. Fig.~\ref{fig:b50_scurve} shows that on the full set of traces (including unseen traces) with the most constrained 50\% L2 + 50\% LLC configuration, \runBS{} slows down only 2.5\% of the 610 single-core traces on fullBW and 7.2\% on limitBW. BertiGO slows down 24.6\% and 14.8\%, and ArchAgent~v2 7.5\% and 11.1\%. At fullBW, \runBS{} outperforms BertiGO and ArchAgent~v2 on 522 and 546 of 610 traces and slows down none of the 359 Google traces. Its robustness advantage is even clearer on four cores. \runBS{} slows down only 14 of 483 4C mixes, against 111 for BertiGO and 136 for ArchAgent~v2, and beats them on 437 and 425 mixes. On 4C Graph mixes it gains +12.1\% and slows down none of the 90 mixes, whereas BertiGO and ArchAgent~v2 lose $-$0.03\% and $-$3.0\% and slow down 42 and 54 of them. Finally, \runBS{} generalizes cleanly to unseen workloads: its lead over ArchAgent~v2 and BertiGO holds or widens on the 479 held-out single-core traces (+4.2\% and +5.5\% on fullBW; +3.7\% and +6.1\% on limitBW), and on the 469 unseen 4C mixes it achieves +12.0\% over the same-size baseline (close to its +14.0\% gain on the 14 searched mixes).

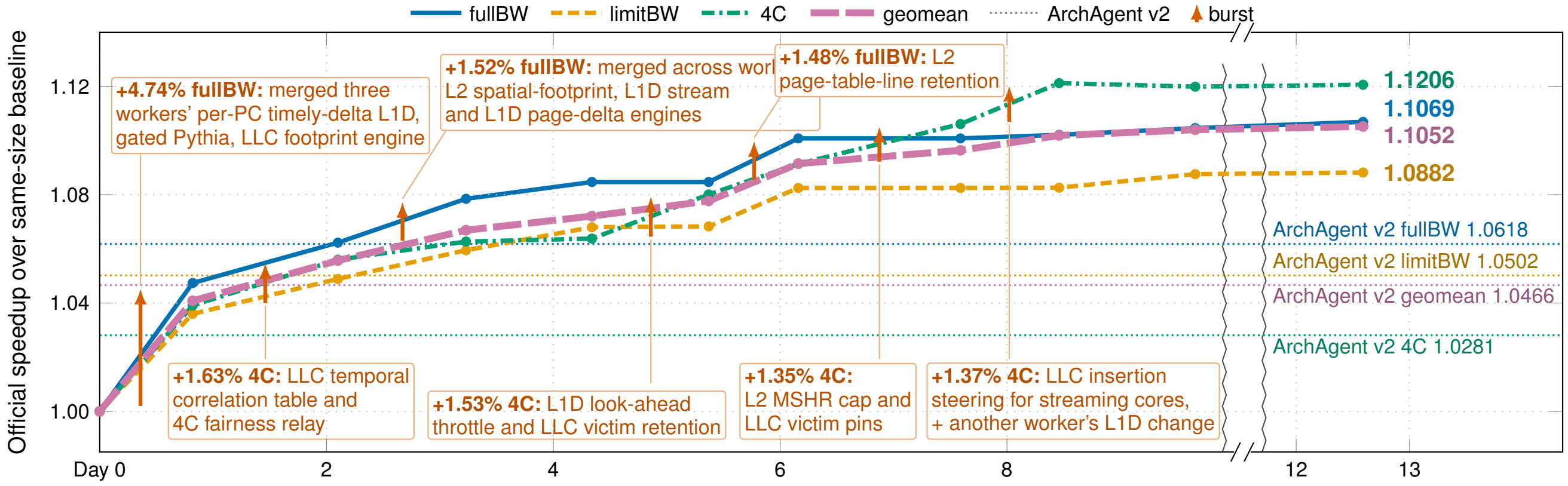
\begin{figure*}[t]
\centering
\definecolor{cFull}{HTML}{0072B2}
\definecolor{cLim}{HTML}{E69F00}
\definecolor{cMC}{HTML}{009E73}
\definecolor{cGeo}{HTML}{CC79A7}
\definecolor{cReg}{HTML}{D55E00}
\definecolor{cNote}{HTML}{555555}
\providecommand{\fann}{\fontfamily{phv}\fontsize{6.44}{7.5}\selectfont}
\providecommand{\flab}{\fontfamily{phv}\fontsize{7.36}{8.6}\selectfont}
\tikzset{ctlimdash/.style={dash pattern=on 3.5pt off 1.8pt}, ctmcdash/.style={dash pattern=on 4pt off 1.5pt on 1pt off 1.5pt},
  ctgeodash/.style={dash pattern=on 9pt off 2pt},
  ctburst/.style={-{Latex[length=1.8mm,width=1.4mm]}, cReg, very thick},
  ctbox/.style={fill=white, draw=cReg!50, rounded corners=1pt, inner sep=1.4pt, text=cReg!85!black}}
\begin{tikzpicture}[font=\fann]
\begin{axis}[name=bvmain, scale only axis, width=\dimexpr\textwidth-1.6cm\relax, height=4.65cm,
  xmin=0, xmax=12.9, ymin=0.985, ymax=1.140,
  x filter/.expression={x>11.7 ? x-1.45 : (x>9.9 ? nan : x)},
  ylabel={Official speedup over same-size baseline}, ylabel style={font=\flab},
  xtick={0,2,4,6,8,10.55,11.55}, xticklabels={Day 0,2,4,6,8,12,13},
  ytick={1.00,1.04,1.08,1.12}, tick label style={font=\fann}, /pgf/number format/assume math mode=true,
  yticklabel style={/pgf/number format/fixed, /pgf/number format/fixed zerofill, /pgf/number format/precision=2},
  grid=major, grid style={dotted, gray!45}, axis on top, clip=false]
\fill[white] (axis cs:9.90,0.985) rectangle (axis cs:10.25,1.140);
\draw[gray!60!black, thin] (axis cs:9.90,0.985) -- ++(1.2pt,3.6pt) -- ++(-1.2pt,3.6pt) -- ++(1.2pt,3.6pt) -- ++(-1.2pt,3.6pt) -- ++(1.2pt,3.6pt) -- ++(-1.2pt,3.6pt) -- ++(1.2pt,3.6pt) -- ++(-1.2pt,3.6pt) -- ++(1.2pt,3.6pt) -- ++(-1.2pt,3.6pt) -- ++(1.2pt,3.6pt) -- ++(-1.2pt,3.6pt) -- ++(1.2pt,3.6pt) -- ++(-1.2pt,3.6pt) -- ++(1.2pt,3.6pt) -- ++(-1.2pt,3.6pt) -- ++(1.2pt,3.6pt) -- ++(-1.2pt,3.6pt) -- ++(1.2pt,3.6pt) -- ++(-1.2pt,3.6pt) -- ++(1.2pt,3.6pt) -- ++(-1.2pt,3.6pt) -- ++(1.2pt,3.6pt) -- ++(-1.2pt,3.6pt) -- ++(1.2pt,3.6pt) -- ++(-1.2pt,3.6pt) -- ++(1.2pt,3.6pt) -- ++(-1.2pt,3.6pt) -- ++(1.2pt,3.6pt) -- ++(-1.2pt,3.6pt) -- ++(1.2pt,3.6pt) -- ++(-1.2pt,3.6pt) -- ++(1.2pt,3.6pt) -- ++(-1.2pt,3.6pt);
\draw[gray!60!black, thin] (axis cs:10.25,0.985) -- ++(1.2pt,3.6pt) -- ++(-1.2pt,3.6pt) -- ++(1.2pt,3.6pt) -- ++(-1.2pt,3.6pt) -- ++(1.2pt,3.6pt) -- ++(-1.2pt,3.6pt) -- ++(1.2pt,3.6pt) -- ++(-1.2pt,3.6pt) -- ++(1.2pt,3.6pt) -- ++(-1.2pt,3.6pt) -- ++(1.2pt,3.6pt) -- ++(-1.2pt,3.6pt) -- ++(1.2pt,3.6pt) -- ++(-1.2pt,3.6pt) -- ++(1.2pt,3.6pt) -- ++(-1.2pt,3.6pt) -- ++(1.2pt,3.6pt) -- ++(-1.2pt,3.6pt) -- ++(1.2pt,3.6pt) -- ++(-1.2pt,3.6pt) -- ++(1.2pt,3.6pt) -- ++(-1.2pt,3.6pt) -- ++(1.2pt,3.6pt) -- ++(-1.2pt,3.6pt) -- ++(1.2pt,3.6pt) -- ++(-1.2pt,3.6pt) -- ++(1.2pt,3.6pt) -- ++(-1.2pt,3.6pt) -- ++(1.2pt,3.6pt) -- ++(-1.2pt,3.6pt) -- ++(1.2pt,3.6pt) -- ++(-1.2pt,3.6pt) -- ++(1.2pt,3.6pt) -- ++(-1.2pt,3.6pt);
\coordinate (bvbrk-b) at (axis cs:10.075,0.985); \coordinate (bvbrk-t) at (axis cs:10.075,1.140);
\draw[cFull, densely dotted, semithick] (axis cs:0,1.0618) -- (axis cs:12.9,1.0618);
\draw[cLim, densely dotted, semithick] (axis cs:0,1.0502) -- (axis cs:12.9,1.0502);
\draw[cGeo, densely dotted, semithick] (axis cs:0,1.0466) -- (axis cs:12.9,1.0466);
\draw[cMC, densely dotted, semithick] (axis cs:0,1.0281) -- (axis cs:12.9,1.0281);
\node[text=cFull!85!black, inner sep=0.6pt, anchor=south west] at (axis cs:10.32,1.0620) {ArchAgent v2 fullBW 1.0618};
\node[text=cLim!70!black, inner sep=0.6pt, anchor=south west] at (axis cs:10.32,1.0504) {ArchAgent v2 limitBW 1.0502};
\node[text=cGeo!70!black, inner sep=0.6pt, anchor=north west] at (axis cs:10.32,1.0464) {ArchAgent v2 geomean 1.0466};
\node[text=cMC!80!black, inner sep=0.6pt, anchor=north west] at (axis cs:10.32,1.0279) {ArchAgent v2 4C 1.0281};
\addplot[cFull, line width=1.4pt] coordinates {(0.00,1.0000) (0.82,1.0474) (2.10,1.0623) (3.23,1.0785) (4.34,1.0847) (5.37,1.0847) (6.16,1.1008) (7.59,1.1008) (8.46,1.1021) (9.66,1.1046) (12.59,1.1069)};
\addplot[cLim, line width=1.4pt, ctlimdash] coordinates {(0.00,1.0000) (0.82,1.0360) (2.10,1.0489) (3.23,1.0595) (4.34,1.0680) (5.37,1.0683) (6.16,1.0825) (7.59,1.0825) (8.46,1.0826) (9.66,1.0876) (12.59,1.0882)};
\addplot[cMC, line width=1.4pt, ctmcdash] coordinates {(0.00,1.0000) (0.82,1.0391) (2.10,1.0560) (3.23,1.0627) (4.34,1.0638) (5.37,1.0801) (6.16,1.0914) (7.59,1.1061) (8.46,1.1212) (9.66,1.1199) (12.59,1.1206)};
\addplot[cGeo, line width=2.2pt, ctgeodash] coordinates {(0.00,1.0000) (0.82,1.0408) (2.10,1.0557) (3.23,1.0669) (4.34,1.0721) (5.37,1.0777) (6.16,1.0915) (7.59,1.0964) (8.46,1.1019) (9.66,1.1040) (12.59,1.1052)};
\addplot[only marks, mark=*, mark size=1.4pt, cFull] coordinates {(0.00,1.0000) (0.82,1.0474) (2.10,1.0623) (3.23,1.0785) (4.34,1.0847) (5.37,1.0847) (6.16,1.1008) (7.59,1.1008) (8.46,1.1021) (9.66,1.1046) (12.59,1.1069)};
\addplot[only marks, mark=*, mark size=1.4pt, cLim] coordinates {(0.00,1.0000) (0.82,1.0360) (2.10,1.0489) (3.23,1.0595) (4.34,1.0680) (5.37,1.0683) (6.16,1.0825) (7.59,1.0825) (8.46,1.0826) (9.66,1.0876) (12.59,1.0882)};
\addplot[only marks, mark=*, mark size=1.4pt, cMC] coordinates {(0.00,1.0000) (0.82,1.0391) (2.10,1.0560) (3.23,1.0627) (4.34,1.0638) (5.37,1.0801) (6.16,1.0914) (7.59,1.1061) (8.46,1.1212) (9.66,1.1199) (12.59,1.1206)};
\addplot[only marks, mark=*, mark size=1.5pt, cGeo] coordinates {(0.00,1.0000) (0.82,1.0408) (2.10,1.0557) (3.23,1.0669) (4.34,1.0721) (5.37,1.0777) (6.16,1.0915) (7.59,1.0964) (8.46,1.1019) (9.66,1.1040) (12.59,1.1052)};
\draw[ctburst] (axis cs:0.36,1.0020) -- (axis cs:0.36,1.0455);
\draw[cReg!50, thin] (axis cs:0.36,1.0465) -- (axis cs:0.36,1.0950);
\node[ctbox, align=left, anchor=south west] at (axis cs:0.10,1.0950)
  {\textbf{+4.74\% fullBW:} merged three\\workers' per-PC timely-delta L1D,\\gated Pythia, LLC footprint engine};
\draw[ctburst] (axis cs:1.46,1.0400) -- (axis cs:1.46,1.0550);
\draw[cReg!50, thin] (axis cs:1.46,1.0390) -- (axis cs:1.46,1.0193);
\node[ctbox, align=left, anchor=south west] at (axis cs:0.60,0.9895)
  {\textbf{+1.63\% 4C:} LLC temporal\\correlation table and\\4C fairness relay};
\draw[ctburst] (axis cs:2.67,1.0630) -- (axis cs:2.67,1.0775);
\draw[cReg!50, thin] (axis cs:2.67,1.0785) -- (axis cs:3.05,1.1035);
\node[ctbox, align=left, anchor=south west] at (axis cs:3.00,1.1035)
  {\textbf{+1.52\% fullBW:} merged across workers:\\L2 spatial-footprint, L1D stream\\and L1D page-delta engines};
\draw[ctburst] (axis cs:4.86,1.0645) -- (axis cs:4.86,1.0795);
\draw[cReg!50, thin] (axis cs:4.86,1.0635) -- (axis cs:4.86,1.0104);
\node[ctbox, align=left, anchor=south west] at (axis cs:2.89,0.9895)
  {\textbf{+1.53\% 4C:} L1D look-ahead\\throttle and LLC victim retention};
\draw[ctburst] (axis cs:5.77,1.0855) -- (axis cs:5.77,1.1000);
\draw[cReg!50, thin] (axis cs:5.77,1.1010) -- (axis cs:6.00,1.1160);
\node[ctbox, align=left, anchor=south west] at (axis cs:5.95,1.1160)
  {\textbf{+1.48\% fullBW:} L2\\page-table-line retention};
\draw[ctburst] (axis cs:6.875,1.0922) -- (axis cs:6.875,1.1053);
\draw[cReg!50, thin] (axis cs:6.875,1.0912) -- (axis cs:6.875,1.0193);
\node[ctbox, align=left, anchor=south west] at (axis cs:5.64,0.9895)
  {\textbf{+1.35\% 4C:}\\L2 MSHR cap and\\LLC victim pins};
\draw[ctburst] (axis cs:8.02,1.1069) -- (axis cs:8.02,1.1204);
\draw[cReg!50, thin] (axis cs:8.02,1.1059) -- (axis cs:8.02,1.0193);
\node[ctbox, align=left, anchor=south west] at (axis cs:7.29,0.9895)
  {\textbf{+1.37\% 4C:} LLC insertion\\steering for streaming cores,\\+ another worker's L1D change};
\node[font=\flab\bfseries, text=cMC!85!black, anchor=west] at (axis cs:11.23,1.1225) {1.1206};
\node[font=\flab\bfseries, text=cFull, anchor=west] at (axis cs:11.23,1.1120) {1.1069};
\node[font=\flab\bfseries, text=cGeo!80!black, anchor=west] at (axis cs:11.23,1.1020) {1.1052};
\node[font=\flab\bfseries, text=cLim!80!black, anchor=west] at (axis cs:11.23,1.0882) {1.0882};
\end{axis}
\foreach \c in {bvbrk-b,bvbrk-t} {\fill[white] ([xshift=-2.6pt,yshift=-2pt]\c) rectangle ([xshift=2.6pt,yshift=2pt]\c);
  \draw[thin] ([xshift=-3.6pt,yshift=-3pt]\c) -- ([xshift=-0.6pt,yshift=3pt]\c); \draw[thin] ([xshift=0.6pt,yshift=-3pt]\c) -- ([xshift=3.6pt,yshift=3pt]\c);}
\node[anchor=south, font=\fann, inner sep=0pt] at ([yshift=2pt]bvmain.north) {%
  \tikz[baseline=-2.5pt]\draw[cFull, line width=1.4pt] (0,0) -- (16pt,0);~fullBW\quad
  \tikz[baseline=-2.5pt]\draw[cLim, line width=1.4pt, ctlimdash] (0,0) -- (16pt,0);~limitBW\quad
  \tikz[baseline=-2.5pt]\draw[cMC, line width=1.4pt, ctmcdash] (0,0) -- (16pt,0);~4C\quad
  \tikz[baseline=-2.5pt]\draw[cGeo, line width=2.2pt, ctgeodash] (0,0) -- (20pt,0);~geomean\quad
  \tikz[baseline=-2.5pt]\draw[gray, densely dotted, semithick] (0,0) -- (16pt,0);~ArchAgent v2\quad
  \tikz[baseline=-2.5pt]\draw[ctburst] (0,-3pt) -- (0,3pt);~burst};
\end{tikzpicture}
\caption{Evolutionary trajectory of \runBS{} on the 50\% L2 + 50\% LLC machine (official speedup over the same-size baseline). Callouts highlight the major step gains and the mechanisms they introduced (the x-axis omits an idle gap on days 9.9--11.7).}
\label{fig:b50_traj}
\end{figure*}

\oldsubsection{How Yardsticks Helped Evolve the Design}
\label{sec:b50found}
\label{sec:yardsticks}
To evolve \runBS{} along the trajectory in Fig.~\ref{fig:b50_traj} without waiting hours per idea on the cluster, workers use custom yardsticks that run 3$\times$ to 65$\times$ faster in wall-clock time than a search-set evaluation. They use them in two complementary ways: as \emph{bottleneck-diagnostic yardsticks} to isolate hardware bottlenecks before writing prefetcher code, and as \emph{candidate-ranking yardsticks} to screen full C++ prefetcher edits before cluster submission.

\begin{table*}[t]
\centering\footnotesize
\caption{Bottleneck-diagnostic yardsticks built by \sysname{} on the 50\% L2 + 50\% LLC machine and the mechanisms it drove.}
\label{tab:b50_bottleneck}
\label{tab:yex}
\begin{tabular}{@{}p{0.09\textwidth}p{0.18\textwidth}p{0.39\textwidth}p{0.27\textwidth}@{}}
\toprule
Level & Diagnostic Yardstick & What It Measures \& Key Finding & Resulting Mechanism (Search Gain) \\
\midrule
\raggedright L1D & \raggedright Miss-latency probe & \raggedright Measures L1D miss latency on 1C vs.\ 4C: per-PC Berti beat page-keyed Berti (+2--3\% 1C, +2.4\% 4C), but its aggressive streams pushed 4C miss latency to 1,000--1,200 cycles (vs.\ $\sim$250 on 1C). & Per-PC timely-delta Berti with a high-latency throttle ($\le 1$ prefetch/trigger above 350 cycles). \\
\addlinespace
\raggedright L2 & \raggedright Per-trace ablation & \raggedright Disables the baseline's L2 Pythia trace-by-trace under limited bandwidth: gained +20.0\% on a SPEC trace and +9.2\% on a Graph trace, but lost $-$9.3\% on BFS (ruling out a static switch). & Online set-dueling gate that dynamically enables or disables Pythia at runtime. \\
\addlinespace
\raggedright L2 & \raggedright Idealized page-walk \& pinning oracle & \raggedright Simulates 0-cycle page walks and pins PTE lines in L2 vs.\ LLC: free walks gained +7--19\% on Google traces, and pinning PTEs in the 1\,MB L2 gained +2.2--8.3\% (vs.\ $\sim$0\% in the LLC). & L2 PTE retention engine (pins cost no extra storage) that infers PTE lines and refreshes them at MRU (+1.24\%). \\
\addlinespace
\raggedright LLC & \raggedright Offline miss-stream profiler & \raggedright Profiles Markov coverage of residual LLC misses: covered 33--43\% on pointer-chasing SPEC traces (2--20\% on Google), while 4C co-runners caused 30\% of a victim's link evictions. & 256\,KB LLC temporal correlator (+0.65\%) with sampled table writes under 4C (+0.46\% 4C). \\
\addlinespace
\raggedright LLC and L1D & \raggedright Single-mix replacement oracle & \raggedright Tests silencing fast cores' L1Ds (+9--17\% on two mixes, $-$27\% on two others) and inserting co-runner LLC fills at lowest priority (lifting the slowest victim core's IPC by +34\%). & Victim-conditional fairness relay (+1.5\% 4C) and borrowed-way victim pins. \\
\bottomrule
\end{tabular}
\end{table*}

\noindent\textbf{Bottleneck-diagnostic yardsticks.} Inside its sandbox, each worker receives the ChampSim source tree, the 131 disclosed traces, and the shipped yardstick template. Because the worker has full write access to its private copy of the simulator, it can construct target-specific yardsticks by instrumenting or modifying its local simulator, such as simulating zero-cycle page walks, pinning cache lines at a chosen level, or adding custom latency probes. These measure how much IPC a bottleneck costs and at which cache level to fix it \emph{before} the worker invests turns in a full C++ prefetcher (Table~\ref{tab:b50_bottleneck}). These local simulator edits never affect cluster evaluation, as the Submission Engine compiles only the submitted prefetcher modules against a clean copy of the simulator (\S\ref{sec:method}). Most of \runBS{}'s key mechanisms originated from such diagnostic experiments designed by the agents themselves, driving the evolutionary trajectory in Fig.~\ref{fig:b50_traj}: early steps established the three-level predictor backbone and initial fairness relay, and later steps alternated between single-core L2 capacity breakthroughs and 4C shared-LLC contention controls. Two examples illustrate how agents used oracle tests to isolate small-cache bottlenecks before writing prefetcher code:
\begin{itemize}
\item \textbf{Diagnosing L2 page-table evictions (+1.24\% search):} Before coding, a worker ran oracle simulations on Google traces showing that free page walks yielded +7--19\% IPC and keeping PTE lines resident in the L2 yielded +6--17\%, proving that leaf PTE lines were thrashing out of both halved caches between page walks. A second oracle pinpointed the right cache level: pinning PTE lines in the shared LLC yielded $\sim$0\%, whereas pinning them in the 1\,MB L2 yielded +2.2--8.3\%. Because the prefetcher API exposes neither page-walk flags nor replacement state, the worker invented the unexplained-fill inference and prefetch-hit MRU refresh mechanism (+1.24\% search gain, plus +0.15\% in a later step that refreshes on every L2 access).
\item \textbf{Diagnosing 4C straggler starvation and LLC victim pinning:} Observing that the 4C score is bottlenecked by the slowest core in each mix, an early worker found that unconditionally silencing fast cores' L1D prefetchers rescued two mixes (+9\% and +17\%) but hurt two others ($-$27\%); conditioning the L1D backoff on a low-IPC, high-MPKI, low-usefulness victim core yielded +1.5\% on the 4C search set. Later, another worker ran a single-mix replacement oracle showing that deprioritizing co-runner LLC fills boosted the victim core's IPC by +34\%, motivating the borrowed-way victim pins (+0.9\% on the 4C grade combined with the L2 MSHR cap); one step later, workers added streaming-core insertion steering (+2.5\% on the 4C grade) without a prior oracle.
\end{itemize}

\begin{table}[t]
\centering\footnotesize
\caption{Speed--fidelity trade-off across candidate-ranking yardsticks in \runBS{}, grouped by trace count.}
\label{tab:y_kinds}
\setlength{\tabcolsep}{3.5pt}
\begin{tabular}{@{}llcc@{}}
\toprule
Trace Counts & Typical Window & Faster By & Ordered Right \\
\midrule
56 (Template) & 2M/8M & $3.0\times$ & 86\% (19/22) \\
16--20 & 1M/4M--3M/10M & $26\times$ & 77\% (55/71) \\
7--15 & 1M/2M--2M/2M & $65\times$ & 57\% (142/250) \\
$\le 6$ & 0M/1M--2M/3M & $59\times$ & 50\% (67/135) \\
\bottomrule
\end{tabular}
\end{table}

\noindent\textbf{Candidate-ranking yardsticks.} Once workers implement a candidate prefetcher, they screen it locally using a candidate-ranking yardstick before submitting to the cluster. Each worker begins with a template that we calibrated offline before search. It runs 56 disclosed single-core traces at fullBW, stratified in proportion to each suite's share of the 131 disclosed traces (9 SPEC17, 18 Google, 13 AI/ML, and 16 Graph), while bounding per-trace runtime so the parallel batch finishes in 14--15 minutes. It also runs 27 of them at limitBW (the traces with the largest baseline IPC drop at 800\,MT/s). All run over a 2M warmup / 8M instruction window, the shortest window in our calibration that preserved rankings across known DPC4 prefetchers. Workers may then customize this template freely, producing 202 distinct trace-set and window configurations across the run. In doing so, workers independently rediscovered sampled simulation, turning both the trace-count and instruction-window knobs to trade simulation time against ranking fidelity. Alongside each search-set evaluation, the Submission Engine re-runs the worker's yardstick on the frozen snapshot outside the sandbox to record whether it orders candidates the same way as the search set (\S\ref{sec:method}). Table~\ref{tab:y_kinds} pools these checks across \runBS{}, comparing median wall-clock speedup over a search-set evaluation against pairwise ordering accuracy on candidates separated by $\ge 0.1\%$. The 56-trace template ($3.0\times$ faster in wall time, $9.6\times$ less simulated work) orders 86\% of pairs correctly (91\% across workers). Moderately pruned 16--20-trace subsets that retain 4M--10M windows ($26\times$ faster, $25\text{--}50\times$ less work) order 77\% correctly. Cutting the trace count below 15 while slashing the window to 1M--3M instructions ($59\text{--}65\times$ faster, bounded by build and launch overhead) degrades agreement to 57\% and then to a 50\% coin flip.

\noindent\textbf{Search-set verification protects final quality.} Even when workers use smaller ($\le 15$-trace) yardsticks for rapid sanity checks during a turn, noisy local proxies cannot hurt the run: a yardstick only filters which candidates a worker submits, while every surviving design is selected strictly on the full search set (131 traces and 14--15 mixes). Across \runBS{}, whenever a working yardstick predicted an improvement, the search set agreed 87\% of the time (60/69), and false positives from overly pruned yardsticks were cleanly rejected by the search set (while resetting workers to the 56-trace template at step boundaries prevented over-pruned yardsticks from carrying over across steps). Similarly, on four cores, where short 1M/4M runs had low correlation with 20M grades because shared-LLC contention requires longer warmup (whereas 2-mix 4C checks at the full 20M window reached Kendall $\tau = 0.71$--$0.78$), scoring all 14--15 4C mixes on the cluster search set prevented single-core yardsticks from trading 4C fairness for single-core speed. Together, Table~\ref{tab:y_kinds} highlights a clear sweet spot for simulator-backed agentic search: moderate pruning (16--56 traces at 4M--10M windows) delivers $3\text{--}26\times$ faster local screening at 77--86\% ranking fidelity, while the cluster search set catches any noisy local estimates.

\noindent\textbf{Value of orchestration over standalone agents.} To isolate the benefit of multi-fleet orchestration and cross-worker merges from the underlying model, we compared 
an orchestrated \sysname{} run on the full-size caches
against three independent, unorchestrated Claude Code agents. All runs started from the baseline prefetchers (\S\ref{sec:background}) and ran for 15 budget days using Claude Opus~5 and the 131 disclosed traces. Even when matched at the same compute budget of 72.7k busy core-hours (the standalone agents' usage at day 13.5), our full-size run already led the best standalone agent on fullBW (6.4\% vs.\ 5.8\%). Without an orchestrator to prune stalled branches or merge complementary discoveries across workers, the best standalone agent plateaued after day~11 at a 4.4\% overall geomean (5.8\% fullBW, 4.3\% limitBW, and 3.0\% 4C; the other two agents reached 2.0\% and 3.1\%), whereas our full-size run reached 6.3\% overall (9.1\% fullBW, 5.9\% limitBW, and 3.9\% 4C).

\oldsubsection{Which Level Matters}
\label{sec:b50levels}
\begin{figure}[t]
\centering
\includegraphics[width=\columnwidth]{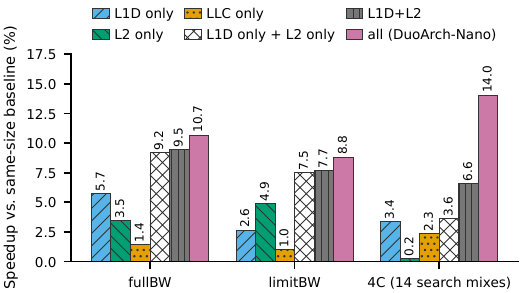}
\caption{Per-level ablation of \runBS{} on the 50\% L2 + 50\% LLC machine (speedup over the same-size baseline).}
\label{fig:b50_levels}
\end{figure}

To quantify how much each cache level contributes, and whether the evolved L1D, L2, and LLC modules act independently or depend on cross-level coordination under halved caches, we evaluate level-by-level combinations of \runBS{} against the same-size baseline on the 50\% L2 + 50\% LLC processor (Fig.~\ref{fig:b50_levels}). The 4C bars use the 14 search mixes, and the L1D + L2 limitBW bar uses 608 of 610 traces because two Graph PageRank traces exceeded the 12-hour job timeout. On single-core workloads, L1D and L2 gains are nearly additive, though their relative importance flips with available memory bandwidth. At fullBW, the rebuilt per-PC L1D drives most of the speedup (5.7\% alone vs.\ 3.5\% for L2 alone). At limitBW, the pollution-gated L2 contributes more alone (4.9\% vs.\ 2.6\%) by curbing inaccurate Pythia traffic and retaining page-table lines in the 1\,MB L2. Combining the evolved L1D and L2 yields 9.5\% on fullBW and 7.7\% on limitBW, closely matching the sum of their isolated gains (9.2\% and 7.5\%), while adding the LLC temporal correlator contributes a further $\sim$1\% to reach 10.7\% and 8.8\%. Under four-core execution (4C), however, the levels exhibit strong cross-level synergy: no individual level achieves more than 3.4\% alone, and combining the L1D and L2 reaches 6.6\%, whereas adding the LLC module more than doubles the four-core speedup to 14.0\% by coordinating cross-level aggressor backoff and shielding victim lines in the contended 6\,MB LLC.

\oldsubsection{Does Scarcity Breed Strength?}
\label{sec:scaling}
\label{sec:ablations}

\begin{figure}[t]
\centering
\includegraphics[width=\columnwidth]{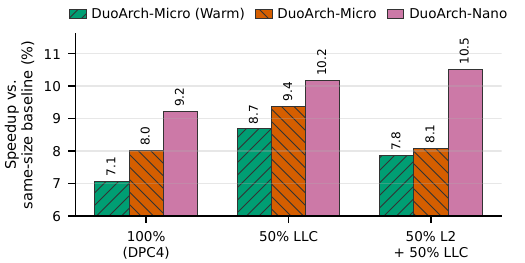}
\caption{Overall official speedup (geomean) of the three small-cache designs over the baseline on the same machine.}
\label{fig:sizes_all}
\end{figure}

To test whether evolving under severe cache scarcity is what made \runBS{} strong, or whether a less constrained small-cache run would have reached the same result, we compare \runBS{} on all three processor configurations against two runs evolved on the 50\% LLC machine (2\,MB L2, 6\,MB shared LLC). \textbf{\runLM{}} was warm-started from an earlier design evolved with the full-size caches~\cite{agarwal2026duodiscover}, and \textbf{\runLS{}} was evolved from the unmodified baseline (official evaluations in Fig.~\ref{fig:sizes_all}).

\noindent\textbf{Cold-starting from the baseline beats warm-starting and matches the baseline with twice the LLC.} First, evolving a small-cache prefetcher does not require a warm-started seed. On the 50\% LLC machine, cold-starting from the unmodified baseline (\runLS{}) nearly catches the warm-started run (\runLM{}) on the search set and edges it out on the Official Evaluation Set (Fig.~\ref{fig:sizes_all}). \runLS{} achieves 9.38\% over the same-size baseline and $-$0.02\% against the full-size baseline (+3.56\% on fullBW, $-$1.09\% on limitBW, and $-$2.42\% on 4C) compared to 8.68\% same-size ($-$0.68\% vs.\ the full-size baseline) for \runLM{}. Even before shrinking the L2, direct evolution from the baseline recovers the overall IPC cost of halving the LLC.

\noindent\textbf{Tighter cache constraints breed stronger prefetchers on every machine size.} Second, rather than over-specializing a prefetcher to small caches, evolving under tighter cache capacity improves overall performance on all three machines (Fig.~\ref{fig:sizes_all}). \runBS{} started from the deepest deficit on the search set ($-$14.6\% vs.\ the full-size baseline), climbing to $-$4.1\% and recovering 72\% of the search-set loss. Yet its design outperforms both 50\% LLC runs overall on every machine. On the both-halved processor, it reaches 10.52\% over the same-size baseline, against 8.08\% for \runLS{} and 7.85\% for \runLM{}. On their own 50\% LLC home turf, it reaches 10.16\% over the same-size baseline and +0.88\% vs.\ the full-size baseline (4C 1.5\% above it), ahead of $-$0.02\% for \runLS{} and $-$0.68\% for \runLM{}. With the full-size caches, the two 50\% LLC runs reach 7.07\% (\runLM{}) and 8.00\% (\runLS{}) against 9.21\% for \runBS{}.

\noindent\textbf{Why tighter caches produce stronger prefetchers.} On a full-size cache (2\,MB L2 and 12\,MB shared LLC), abundant SRAM masks prefetch inefficiency: moderately inaccurate prefetches or uncoordinated four-core bursts are absorbed by spare cache ways without triggering thrashing, allowing search to plateau in a local optimum. Shrinking the caches strips away that cushion, turning every wasted cache slot or unthrottled aggressor core into an immediate capacity eviction and 4C harmonic-mean penalty. Consequently, decomposing each step's gain in the two 50\% LLC runs shows that, once the base predictors were in place, most of the gain came from the four-core score, that is, from fixing inter-core contention rather than boosting single-core aggressiveness.

\noindent\textbf{What the 50\% LLC runs built.} Starting from the baseline without design hints, \runLS{} (like \runBS{}, \S\ref{sec:both50design}) first converged on known predictor families at all three levels (+6.18\% search gain in its first step): a per-IP timely-delta L1D prefetcher, an L2 Pythia behind a perceptron usefulness filter, and a 256\,KB LLC temporal correlation table with spatial footprints~\cite{triage,triangel,sms,bingo}, later adding store-stream reconstruction. After that initial step, 56--100\% of \runLS{}'s step gains and 83--100\% of \runLM{}'s early step gains came from the four-core score:
\begin{itemize}
\item \textbf{Code-pressure demotion and per-core LIFO queues (\runLS):} Because instruction fetches never invoke the prefetcher hook, sampled L2 sets track a data-only shadow and infer an instruction-line eviction whenever a data fill evicts an unseen address, demoting weak data prefetches to LLC-only under code pressure~\cite{srinath2007feedback,emissary} (part of a +1.00\% search step that also added store-stream reconstruction, after a +1.91\% fairness-tuning step). At the shared LLC, a later step (+0.54\% search overall and +0.92\% on the 4C search score, together with a page-walk line keeper) replaced FIFO emission on 4C with per-core newest-first (LIFO) stacks drained round-robin when the LLC prefetch queue holds fewer than 4 entries, preventing bursts from starving co-runners.
\item \textbf{Three-level straggler-protecting governors (\runLM):} Recognizing that the 4C harmonic-mean score is dominated by the slowest core, \runLM{} kept the starting design intact and added coordinated QoS governors~\cite{coordthrottle} at all three levels, gaining +0.5\% 4C at the LLC, +2.3\% 4C at the L1D (capping look-ahead for the 3 fastest cores when the leader/straggler retired-instruction ratio exceeds 24/8), and +1.3\% 4C at the L2, with whole-step gains of +0.75\% and +0.46\%.
\end{itemize}

\noindent\textbf{Why \runBS{}'s mechanisms go further.} While all three small-cache runs retain page-table lines~\cite{csalt} (reaching similar single-core L2 page-walk misses of 0.51--0.54\,PKI at full size) and throttle four-core contention, they differ in how they manage the shared LLC. Specifically, \runLM{}'s governors only slow fast cores whenever they pull ahead, and \runLS{} only throttles prefetch aggressiveness and orders each core's LLC prefetches newest-first; neither directly protects the slow victim core's resident LLC lines. Forced to operate with a halved 1\,MB L2 in front of the 6\,MB shared LLC, \runBS{} built mechanisms akin to cache partitioning and insertion control~\cite{ucp,pipp} that protect those lines directly without modifying replacement hardware (\S\ref{sec:both50design}). In particular, it pins the slowest core's LLC hits in two ways borrowed from the correlation table via prefetch-hit refreshes, steers DRRIP insertion so streaming cores' fills are evicted first, and restricts noisy co-runners' writes to the shared correlation table. As a result, almost all of \runBS{}'s full-size advantage over \runLS{} comes on four cores. On the full-size machine, \runBS{} beats \runLS{} on 327 of 483 4C mixes, lifting 4C SPEC17 from +13.9\% to +21.6\% and turning 4C Graph from a $-$1.9\% loss into a +6.1\% gain. At the same time, it incurs 14\% fewer 4C LLC demand misses and issues 17\% fewer DRAM requests per instruction (11\% fewer than the baseline), with per-mix IPC gains closely tracking LLC miss reductions (Spearman $\rho = 0.69$).

\oldsubsection{Where the Gap Remains}
\begin{figure*}[t]
\centering
\includegraphics[width=0.95\textwidth]{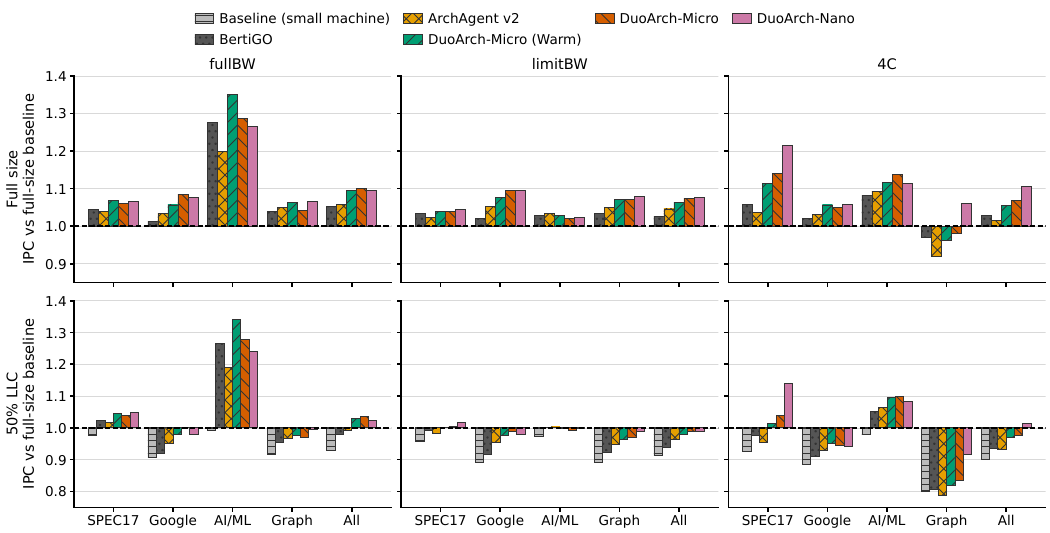}
\caption{Official IPC per suite against the full-size baseline 
at fullBW, limitBW and 4C.}
\label{fig:suite3}
\end{figure*}

Fig.~\ref{fig:suite3} breaks the full-size and 50\% LLC results down by suite. Why does evolution completely close the overall capacity gap at 50\% LLC, yet remain $-$6.20\% below the full-size baseline when both caches are halved? Breaking the official results down by workload suite reveals a sharp divide between regular workloads (SPEC17 and AI/ML) and irregular or large-footprint workloads (Graph and Google). Halving the LLC converts working-set reuse that previously fit on-chip into capacity misses whose irregular access patterns exceed the coverage of a 256\,KB on-chip prefetcher: across traces, the IPC loss from shrinking the cache closely tracks the rise in LLC misses (Spearman $\rho = 0.92$--$0.94$ within Google and Graph). On single-core runs at 50\% LLC, the two 50\% LLC runs' prefetchers erase 100\% of the extra capacity misses on SPEC17 and AI/ML, 46--82\% on Google (allowing \runLS{} to match the full-size baseline at +0.2\% on fullBW Google), and 37--55\% on Graph. Under four-core execution (4C), however, co-running cores competing for the halved shared LLC amplify Graph's extra capacity misses eightfold and nearly double DRAM traffic on an already queue-bound memory bus; with both caches halved (not shown in Fig.~\ref{fig:suite3}), \runBS{} still finishes at $-$17.0\% on 4C Graph and $-$13.0\% on 4C Google relative to the full-size baseline.

\needspace{3\baselineskip}\oldsection{Related Work}
\label{sec:related}

\noindent\textbf{Data prefetchers.} Spatial prefetchers learn which lines of a region a program touches, from region
footprints~\cite{sms,bingo} or instruction-pointer classes~\cite{ipcp}.
Offset and delta prefetchers learn a distance to prefetch at, globally~\cite{bop}, per load instruction~\cite{navarro2022berti}, or as within-page delta signatures followed speculatively with lookahead~\cite{spp};
Caerus~\cite{caerus}, the most recent, learns several global offsets and filters them per load instruction, reporting a 1.5\% lead over BertiGO in its own SPECspeed 2017 setup (when we port the authors' code to DPC4 as an L1D-only replacement for Berti, removing its private tag-array filter per DPC4 rules, it gains 1.0\% overall over the baseline). Temporal prefetchers replay
recorded miss sequences~\cite{isb,triage,triangel}, and Pythia~\cite{bera2021pythia} chooses prefetch actions by online reinforcement
learning. Several of these designs won the Data Prefetching Championship, which compares prefetchers in
ChampSim~\cite{gober2022championship} under fixed storage budgets: a tuned BOP won DPC-2, IPCP won DPC-3, and BertiGO~\cite{singh2026pushing}, built on Berti, won
DPC4~\cite{dpc4}, whose limitBW and multi-core configurations penalize bandwidth-hungry designs. 

\noindent\textbf{Prefetch throttling, shared-cache management, and cache sizing.} To prevent aggressive prefetching from saturating memory bandwidth or polluting caches, prior work throttles prefetch degree using runtime accuracy, pollution, and bandwidth feedback~\cite{srinath2007feedback,coordthrottle} or schedules prefetches at the memory controller~\cite{lee2008prefetch}. Other proposals modify cache replacement and partitioning to curb prefetch pollution~\cite{wu2011pacman,seshadri2015mitigating}, partition or pseudo-partition shared LLC capacity across cores~\cite{ucp,pipp}, retain instruction lines~\cite{emissary}, or partition cache space between data and translations~\cite{csalt}, while studies such as Pythia~\cite{bera2021pythia} evaluate fixed prefetchers across LLC capacities. Rather than modifying replacement hardware or only evaluating designs on smaller caches, \runBS{} operates under unmodified DRRIP~\cite{jaleel2010rrip} through the standard prefetcher interface, and we find that \emph{evolving} prefetchers under cache scarcity produced stronger designs overall at all three cache sizes in our runs (\S\ref{sec:scaling}).

\noindent\textbf{LLM agents for hardware design.} Prior LLM-driven architecture work either targets other structures, such as cache replacement~\cite{gupta2026archagent, mhapsekar2026cacheforge}, parameter-level design-space exploration~\cite{xiong2026microevo, fu2025gem5}, or evolves single-level prefetchers outside the DPC4 setup without the code~\cite{kypriotis2026agentic}. We therefore compare against ArchAgent~v2~\cite{gonzalez2026archagentv2}, the state-of-the-art agentic system on DPC4, which builds on its predecessors~\cite{novikov2025alphaevolve, gupta2026archagent} and evolves DPC4 prefetchers a cache level at a time.

\section{Conclusion}
\label{sec:conclusion}

We presented \textbf{\sysname{}}, a multi-fidelity autonomous discovery system that overcomes cycle-level simulation bottlenecks to co-evolve three-level data prefetchers. DuoArchitect discovered \textbf{\runBS{}}, an intelligent prefetcher evolved under halved L2 and LLC capacities. Evaluated on DPC4 benchmarks, DuoArch-Nano achieves a 9.2\% overall speedup with full-size caches, more than doubling the gains of BertiGO (3.5\%) and ArchAgent v2 (3.9\%). When half the LLC is removed (1.5 MB/core), DuoArch-Nano erases the baseline's 8.6\% slowdown and outperforms the full-cache baseline (+0.88\%), demonstrating that microarchitectural intelligence can directly substitute for physical cache capacity. 

Reclaiming this cache silicon enables a direct boost to compute density. For instance, in an AMD Zen 4 core-complex (where the 32 MB L3 occupies 44\% of the eight-core complex), halving the per-core LLC frees enough silicon area for roughly 20\% more physical cores. Replacing passive cache capacity with active, prefetcher-driven intelligence allows hardware designers to shift the area-performance pareto-frontier beyond technology scaling limits.

More broadly, these results point to a paradigm shift in agentic computer architecture. Rather than serving as design-automation assistants, we show that LLM-based agents can navigate  complex, multi-dimensional trade-offs that are traditionally too massive for human engineers to reason about.  By co-evolving both policies and their own multi-fidelity simulation environments, agentic frameworks enable the replacement of bulky, scaling-stalled physical structures with intelligent software-driven microarchitectures.

\begin{acks}
Gemini 4 Neon was used to assist with writing, plotting scripts, and number checks; the authors reviewed all content and take full responsibility for it.
\end{acks}

\newpage
\let\section\oldsection %
\balance
\bibliographystyle{ACM-Reference-Format}
\bibliography{refs}

\end{document}